# Recognizing Extended Spatiotemporal Expressions by Actively Trained Average Perceptron Ensembles

Wei Zhang
Carnegie Mellon University
5000 Forbes Avenue
Pittsburgh PA 15213, USA
wynnzh@gmail.com

Yang Yu
IBM Watson
550 King Street
Littleton MA 01460, USA
yu@us.ibm.com

Osho Gupta
Indian Institute of Technology
Varanasi, India
osho.gupta.ece11@iitbhu.ac.in

Judith Gelernter
Carnegie Mellon University
5000 Forbes Avenue
Pittsburgh, PA 15213
gelern@cs.cmu.edu

## ABSTRACT

Precise geocoding and time normalization for text requires that location and time phrases be identified. Many state-of-the-art geoparsers and temporal parsers suffer from low recall. Categories commonly missed by parsers are: nouns used in a non-spatiotemporal sense, adjectival and adverbial phrases, prepositional phrases, and numerical phrases. We collected and annotated data set by querying commercial web searches API with such spatiotemporal expressions as were missed by state-of-the-art parsers. Due to the high cost of sentence annotation, active learning was used to label training data, and a new strategy was designed to better select training examples to reduce labeling cost. For the learning algorithm, we applied an average perceptron trained *Featurized Hidden Markov Model* (FHMM). Five FHMM instances were used to create an ensemble, with the output phrase selected by voting. Our ensemble model was tested on a range of sequential labeling tasks, and has shown competitive performance.

Our contributions include (1) an new dataset annotated with named entities and expanded spatiotemporal expressions; (2) a comparison of inference algorithms for ensemble models showing the superior accuracy of Belief Propagation over Viterbi Decoding; (3) a new example re-weighting method for active ensemble learning that "memorizes" the latest examples trained; (4) a spatiotemporal parser that jointly recognizes expanded spatiotemporal expressions as well as named entities.[1]

## Categories and Subject Descriptors
D.3.3 [**Artificial Intelligence**]: Natural Language Processing – *language parsing and understanding, text analysis*

## General Terms
Algorithms

## 1. INTRODUCTION
We consider the problem of recognizing unnamed location and time phrases to ease downstream applications such as geocoding [45] and time normalization [2]. A wider range of expressions parsed could benefit *question and answering* [11][9][31], *web querying toward the semantic web* [43], *information retrieval* as from news messages [32].

**Data.** The types of location and time expressions that parsers miss are the types not included in standard annotated data sets. So we were forced to create our own labelled data because given location and time tags in most corpora are named entities. To create our corpus, we sampled sentences from the web that include seed phrases of the sort that preliminary experiments had shown are not found by geoparsers and temporal parsers. False positives were also collected by search engine. We used active learning to find sentences that would provide the most information gain for training. These 1701 unlabeled training, and 450 manually labeled testing sentences make up our *Expanded Spatiotemporal Data* (**EST**).

**Model.** Labeling words in sentences automatically is done sequence labeling models such as Hidden Markov Models (HMM) [28] [46] and Conditional Random Fields (CRF) [16], Average Perceptron [3], Max-Margin Markov Networks [39], SVM$^{struct}$ [42], SEARN algorithm [5], Max Entropy Markov Models [20] and so on. We are interested in Average Perceptron due to its high accuracy and ease of overfitting control by averaging model parameters. We used an average perceptron algorithm for training a Featurized HMM (an extension of HMM with features, see Section 3). To further reduce overfitting and prediction variance, we used FHMM ensembles, specifically, bagging [6] of FHMM, with the correct result being found by voting among results of the individual models. We also experimented with two different ensemble inference methods called *Best Viterbi Sequence* and *Best BP Sequence* which are based on *Viterbi Decoding* and *Belief Propagation* (BP) respectively.

**Evaluation.** We use commonly used named entity recognition (NER) data sets and a chunking data set to demonstrate the viability of FHMM model in comparison to other methods before running our own EST data.

---
[1] Source at https://github.com/weizh/LearnSpatioTemporal and https://github.com/weizh/PerceptronFHMM

We used five sub-datasets from the OntoNotes[2] news data that others had published NER results, the CoNLL 2000 chunking dataset[3], and the NLPBA biomedical NER dataset[4]. Our single FHMM model performed competitively with others on NER tasks and chunking task on those data sets. Thus we could confirm that FHMM is a reliable building block for an ensemble model. We found also that the Belief Propagation works in concert with the ensemble FHMM better than does the Viterbi, and that our active sampling method achieved higher results overall than would a random sample of the data have achieved.

**Contributions**
(1) A general English Expanded Spatiotemporal Dataset, the EST corpus is created. We used seeds to find sentences on the web, and used active learning to find and tag the examples. (2) We adapted single model inference algorithms to ensembles, and found out that *Best Belief Propagation Sequence* (Belief Propagation variant) won over *Best Viterbi Decoding Sequence* (Viterbi Decoding variant) for ensembles to a large margin, especially for ensembles. This conclusion also guides the selection of inference algorithm for other sequence labeling models as well. (3) A new training example re-weighting method is proposed to reduce the variance of the ensemble created in active learning process, to eventually speed up the active learning process. (4) A spatiotemporal parser that jointly recognizes expanded spatiotemporal expressions as well as named entities.[5]

In section 2, we survey the related work, section 3 we elaborate on our data collection method. Section 4 introduces the average perceptron trained FHMM model, followed by the FHMM ensemble in Section 5. Active Learning method for FHMM ensemble is explained in Section 6. At last, we show experiment results on FHMM, FHMM ensembles, and Active Learning of FHMM ensembles in Section 7.

## 2. RELATED WORK

*No single model for location and time yet.* We join the parsing of time and locations in one model (one FHMM ensemble), because we observed linguistic parallels between spatial and temporal expressions, as is shown in Table 2. We believe that so doing could help distinguish between meanings of words that could be used just as frequently in a temporal as in a location sense.

*What does a parser typically find, and why?* Geoparsers find nouns – specifically, location words (or toponyms). Temporal parsers find months, dates and holidays. Phrases (that includes valuable modifiers to time and locations) are not always included if ever. The PETAR system ([18] that works on tweets is among the first to change this.

*State-of-the-art spatiotemporal parser accuracy* To get a general idea of the precision and recall of existing temporal parsers, we wrote 79 sentences[6] (we use the words "*sentences*", "*instances*" and "*examples*" interchangeably throughout this paper) using specific and non-specific temporal words and phrases that covers a variety of cases, which is a decent benchmark for evaluating

---
[2] https://catalog.ldc.upenn.edu/LDC2011T03
[3] http://www.cnts.ua.ac.be/conll2000/chunking/
[4] http://www.nactem.ac.uk/tsujii/GENIA/ERtask/report.html
[5] Code at https://github.com/weizh/LearnSpatioTemporal and https://github.com/weizh/PerceptronFHMM

[6] Data available at https://github.com/weizh/PerceptronFHMM

---

**Table 1. Temporal tagger accuracy based on 79 sentences**

|  | Precision | Recall | F1 |
|---:|---|---|---|
| Heideltime | 0.827 | 0.409 | 0.547 |
| Illinois extractor | 0.866 | 0.396 | 0.544 |
| GUTime | 0.786 | 0.36 | 0.494 |
| SUTime [2] | 0.874 | 0.677 | 0.763 |
| Stemptag (IntelliGIS) | 0.863 | 0.616 | 0.719 |
| Gmail (Google) | 0.818 | 0.109 | 0.194 |
| Mac OS X (Apple) | 0.833 | 0.122 | 0.213 |
| TIPSem (Univ Alicante) | 1 | 0.25 | 0.4 |

**Table 2. Linguistic parallels between place and time**

|  | Time phrase | Location Phrase |
|---|---|---|
| **Numerical** | Quarter to three | Quarter mile |
|  | Three hours | Three stops on bus |
|  | 25 minutes | 25 light years |
| **Adj/adv** | Quite a long time | Quite close |
|  | Every Tuesday | Every kilometer |
|  | Far from the end | Far away |
| **Prepositions** | Near dinner time | Near the exit |
|  | At last | At the door |
|  | Over an hour | Over the hill |

spatiotemporal parsers. We ran the sentences through temporal taggers in Table 1. For geoparsing (method to find location entities) results, see [45].

*No corpora tagged with spatiotemporal phrases.* Geocoding research that incorporates natural language understanding of adjectival or prepositional phrases has been termed "spatial relations" [44]. Tags for such expressions were defined only recently.[7]

**Ensemble methods** Many temporal parsers are based on heuristic methods, and the geoparsers are based on classifiers with knowledge bases (a world gazetteer). Ensemble methods are generally reserved for problems of higher complexity. [22] and [23] studied ensemble classifiers for continuous values, [33] for gesture recognition, [25] for biomedical data, and [13] on object recognition for computer vision. [19] designed ensembles that are heterogeneous rather than homogeneous.

**Active Learning for Ensembles.** Active learning for ensembles is little studied beyond [22] and [23]. However, active learning for sequence labeling models has been investigated extensively [4] [36] [21].

## 3. Web DATA for spatiotemporal corpus (EST)

*Necessity of creating EST Dataset.* Among all the data sets that we investigated (CoNLL 2003[8], OntoNotes, NLPBA, etc.), the occurrence of phrase types in Table 2 are rare, which will undermine our method due to the training data insufficiency. Thus, we made *Expanded Spatiotemporal Dataset* (EST Dataset) to include examples of spatiotemporal expressions that FHMM ensembles learn to recognize. We used the Microsoft Search API to automatically collect sentences from the web using seed expressions (Figure 1) of the sort missed by state-of-the-art temporal parsers and geoparsers.

---
[7] http://timexportal.wikidot.com/corpora-tides
[8] http://www.cnts.ua.ac.be/conll2003/ner/

**Table 3. Entity type definition**

| Type | Type Definition | Examples |
|---|---|---|
| L | Locations | "United States", "Boston" |
| D | Date, time | "July 7th, 2014", 2am |
| **G** | **Spatial terms** | **"west of", "close to"** |
| **T** | **Temporal terms** | **"Hours before", "during", ..** |
| O | Organization | "WTO", "Coca cola" |
| P | Person | "George Washington" |
| ST | Street, roads, corners | "Fifth street" |
| B | Building | Empire State, Taj Mahal |
| W | Web links and urls | "www.google.com" |
| UL | Unnamed location | "Several states" |
| US | Unnamed street | "The street, the road" |
| UB | Unnamed buildings | "The house, the building" |
| E | Event | "Boston Marathon" |
| [O] | Words that do not belong to the above types. | |

*across from, at once, at the end, at the same time, between, classic, close to, contemporary, early, eastward, every mile, ever since, every time, fall, far away, far from, far longer, first week in, formerly, fortnight, frequently, half of an hour, homeward, in the back, into the distance, last kilometer, last moment, long distance, long time, miles before (the station, intersection), minutes before (closing, sunset, the hour, the match), modern quick, quite close, quite late, rarely, right then, right there, second Sabbath, second weekend in, several blocks (ahead, away), since the day, straight away, third weekend in, toward the, very close, very late, very near, walking distance, without delay*

**Figure 1. Sample of spatiotemporal keywords used to search Microsoft Search API to collect sentences**

***Data collection.*** Figure 1 shows some seed words used to search the Microsoft search API to make the training dataset. Keyword search results might be biased, but the advantages are that this is an efficient way to collect relevant data. For each returned list of results, we selected the top N snippets. We observed the results, and found that the top results were highly biased toward named entities (because of search engine optimization). For instance, "*The Day After Tomorrow*" is a movie name instead of a temporal phrase. To keep the dataset from leaning toward such edge cases, we used N=50, so that more sentences with the general sense for time or location words could be included. We uniformly sampled 2151 sentences from all of those we collected. For evaluation purposes, we randomly split the dataset into 80% training and held 20% for testing. We manually tagged the 450 test sentences according to the tag set in Table 3 and we left the remaining 1701 training sentences untagged for active learning.

***Annotations.*** We created the tag set shown in Table 3. In addition to traditional location entity types such as **L**, we also added the annotation **B** and **ST**, for "Building", "Street" for fine-grained named entities, **UL**, **US**, **UB** for unnamed locations to distinguish them from named ones. We tagged all the words in a time phrase as ***T*** that are not named entities, and all location words as ***G*** that are not location named entities. For example, given the letter-tags in Table 3, *"west of Boston"* is tagged as (***G G L***)**,** and *"3 days before National Day"* as (***T T T E E***)**.** Notice that what is annotated ***GG*** is actually a geo-word plus a preposition. The annotation of the preposition as **T** is exactly what we need in the training data, so that "of" is included as part of the named entity to be used to connect "west" with "Boston" to more precisely target the location or range.

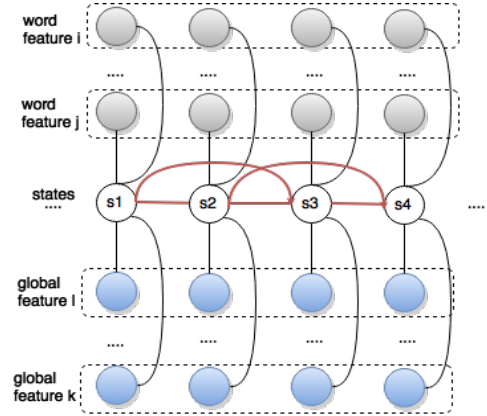

**Figure 2. Structure of Featurized FMM**

---

*Algorithm* **1. The Perceptron training algorithm**

Initialize:
Constant $K$ = total number of features
Constant $F$ = total number of feature values
All weights $a_k \in R^{K \times F}$ with zero, sum of weights $A_k$ with zero.
1 f*or each* iteration $t \in [1, T]$, and training example $e_i$, $i \in [1, N]$:
2     $z(e_i) = FHMM.predict(e_i)$
3     *for each* word position *p*:
4         *for each* feature k*:*
5             $\alpha_k\left(z(e_{ip})\right) = \alpha_k\left(z(e_{ip})\right) - 1;$
5             $\alpha_k\left(y(e_{ip})\right) = \alpha_k\left(y(e_{ip})\right) + 1;$
7 store a snapshot of $\alpha$ into $A_k$
8 output $A_k/(T \times N)$

---

## 4. THE AVERAGE PERCEPTRON TRAINING OF FHMM

We introduce the Featurized Hidden Markov Models (FHMM) (see Figure 2), the building block of our FHMM ensemble. The model is trained with the average perceptron algorithm [3], which resembles a single linear perceptron in neural networks.

Since we apply perceptron training, the model turned from Bayesian model into a discriminative model and model bias is reduced. Thus, there is no arrow for the edges in Figure 2. The gray nodes labeled 'word feature j' define the *j*-th word level feature function for each word in example *S*, such as the shape, part-of-speech or suffix that does not depend on the context. The blue nodes labeled 'global feature k' under the states layer represent the *k*-th global feature function defined on context of a word. In the middle 'states' layer, each state (denoted by *s*) represents a label in Table 3, and the line between states is the second order transition feature. The features are all attached to weights, denoted as $\alpha_k$, where *k* indexes feature functions. The probability of a training example is proportional to the linear combination of its feature values. See [3] for details.

The perceptron algorithm updates feature weights $\alpha_k$ iteratively by observing the errors made by the current parameter setting and adjusts the parameter according to errors (See Algorithm 1). This idea is similar to the back-propagation algorithm used to train neural networks. Averaging parameters in the perceptron can help reduce FHMM overfitting.

**Features.** Table 4 shows how we define features used for the FHMM according to different datasets. We need to use different

**Table 4: Different feature configurations for different tasks. [-2,2] in first column means the feature is extracted within token window of 5 for each word.**

| Feature | Chunk | NLPBA NER | Onto Notes NER | EST NER+ unnamed |
|---|---|---|---|---|
| Current word | y | y | y | y |
| Current word lemma | y | y | y | y |
| Current word part of speech | y |  | y |  |
| Current word suffix | y | y | y | y |
| Current word form+suffix | y | y | y | y |
| Current word prefix + suffix | y |  | y |  |
| Current word POS + suffix | y |  | y |  |
| Lemma in window [-2,2] | y | y | y | y |
| Word form [-2,2] |  | y |  | y |
| Part-of-speech [-2,2] | y |  |  |  |
| Word suffix [-2,2] | y | y |  | y |
| Named entity tag [-2,2] |  |  |  | y |
| Part-of-speech bigram [-2,2] |  |  | y |  |
| Suffix bigram [-2,2] |  | y |  |  |
| Word form bigram [-2,2] |  | y | y | y |

**Algorithm 2. General Query-by-bagging algorithms for sequence labeling ensembles**

$T$: set of training examples, initialized with $I$ examples
$U$: set of unlabeled examples
*Ensemble*: the Ensemble learning algorithm
*Base*: The base sequential labeling algorithm
*Utility*: The utility function defined for unlabeled training example
1 **repeat** t times:
2 Generate *Ensemble* = learn_Ensemble( *Base, T*)
3 **for each** example *e* in U:
$util_e$ = *Utility*(*Ensemble, e*)
4 Sort U by $util_e$, select top $K$ examples *A* in *U*
5    Label examples in *A*, move them from *T* to *U*.
6 **Return** *Ensemble*

**Algorithm 3. Selection by sentence re-weighting**

Input: Training data $T_t$ and corresponding weight vector $W_t$ of iteration t
$\alpha$: Small value used for sample rate decay
$I$: The new example to add to training data $T_t$
1 **for each** example-weight pair$(T_{it}, W_{it})$ in $(T_t, W_t)$, **do**
2 $w_{it} = \max(1 - \alpha(t-1), r)$
3 $W_{t+1} = W_t \oplus 1, T_{t+1} = T_t \cup I$
4 return$(T_{t+1}, W_{t+1})$

feature for different sets because each dataset contains different annotations. For example, the NLPBA corpus does not use the Part-of-Speech (POS) tags, so the part of speech feature is not included. Feature engineering is important because perceptron is a single neuron that is linear with respect to its input. Feature engineering could help the model to be "non-linear" to better fit the NER problem. In Table 4, the features defined in context window [-2,2] are the "global features" that increase non-linearity.

**Overfitting**. Many machine learning algorithms have a high risk of overfitting, especially discriminative models that are non-Bayesian. Several general techniques can be used to reduce overfitting in optimization, such as early stopping [27], L2 regularization [7], Lasso regression [40], cross validation [14], and feature selection. But in a perceptron algorithm, *parameter averaging* is a simple but effective technique used to control overfitting [3]. Additionally, we apply the ensemble learning technique to reduce the variance of the model created, so that overfitting can be further controlled. See section 5 for details.

**Inference.** We compared two inference (decoding) algorithms: *N-best Viterbi decoding* (or *Viterbi decoding* for short) [10], and the *Belief Propagation* (BP) algorithm, or the sum-product message passing algorithm [26], [30]. Viterbi decoding can be used when exact inference is feasible, and BP could also be used for exact inference, but can also give approximate results when the joint probability is impossible to compute. The BP algorithm is widely used in graphical models for exact inference in tree graphs [15] and for approximate inference over loopy graphs [24]. Recent work tries to combine the two to achieve the power of both [17]. In our work, we compare Viterbi decoding and BP algorithm in the setting of sequence labeling ensembles. *Viterbi decoding* generates the probability for each decoding sequence, whereas BP uses forward-backward algorithm to predict each token individually [28]. We noticed that for **single** FHMM model, the inference accuracy of BP is generally worse than *Viterbi decoding*, especially when the dataset is large. However, it is different for ensemble models. We will come back to this point in a moment.

## 5. BAG OF AVERAGE PERCEPTRONS – ENSEMBLE FHMM

Ensemble learning is a meta-learning technique to reduce variance of the training model [1]. The bagging algorithm [1] as a type of ensemble learning algorithm, creates a bag of models (same learning functions) using a subset of training data that are sampled with replacements from the full data. The number of FHMM created is heuristically selected. When the training data size is small, especially during the first several iterations in an active learning procedure, a FHMM model can have high variance on its output (common to other training algorithms as well) due to overfitting, which in turn can lead to imprecise prediction of the next informative sentence to include. To solve this problem, we experiment with candidate inference algorithms so that better predictions can be made and based on the utility function to choose the next instance, in hopes of optimizing selected instances faster.

Due to the change from using single FHMM to an ensemble of FHMMs, strategies are needed to combine the results from FHMMs. We designed two inference algorithms for the ensemble based on the Viterbi Decoding and BP algorithm respectively. The *Best Viterbi Sequence (***BVS***)* strategy is based on *Viterbi decoding* results from each FHMM member. Formally, assume $M = \{M_1, ..., M_k\}$ is an ensemble of $k$ FHMMs. The total collection of unique output sequences from the ensemble is denoted as $S = \{S_1, ..., S_L\}$, coming from all $n$-best sequences of all FHMM models. So the best $S_j$ given the ensemble $M$ is described by

$$S^{best} = argmax_j P_M(S_j) = argmax_j \sum_k P_{M_k}(S_j)$$

Where $P_{M_k}(S_j)$ is the probability for $M_k$ to predict sequence $S_j$ as output. This probability is the aggregated probability from all the models in ensemble for $S_j$. And, the sequence that has the best aggregated probability is the best output $S^{best}$ for ensemble.

By contrast, the *Best BP Sequence* (**BPS**) that corresponds to the BP algorithm for a single model is the aggregation of individual BP predictions. Formally, we define $w_{ij}$ to be the word $j$ in example $e_i$. The *l*-th label for $w_{ij}$ in label sequence $s_{ij}$ is $s_{ij}^l$. The best sequence is

$$s_{ij}^{best} = argmax_l\, P_M(s_{ij}^l\,|e_i) = argmax_l \sum_k P_{M_k}(s_{ij}^l\,|e_i)$$

Where each $P_{M_k}(s_{ij}^l\,|e_i)$ is generated by the BP algorithm from model $M_k$. It shows that the BP decoding result for token $j$ in example $i$ is the aggregated probability from the output of all the models. The best label for token $j$ is the one that gives the best aggregated score. The best decoding sequence is just the concatenation of labels from each individual token. Compared to the *Best Viterbi Sequence*, where the relation among predictions is preserved, the *Best BP Sequence* does not force the coupling of predictions between tokens. Each individual prediction can change freely given the individual output. We observed from our experiment that when a single model is used and the data is abundant, the *Best Viterbi Sequence* generates better result. However, when ensembles of models are used in the first several iterations, the *Best BP Sequence* performs better.

# 6. ACTIVE LEARNING FOR FHMM ENSEMBLES

Active learning for classifier ensembles has been studied on classification problems [22] and [23]. Active learning for sequence labeling models was extensively investigated in [4] [36] [21]. We are the first to use active learning technique to train an ensemble of sequence labeling models. Active learning techniques is used to reduce labeling cost by choosing only the most effective training instances, to achieve a classifier comparable to a model trained with the full dataset. Labeling training examples can be costly.

The FHMM ensemble during active learning produces output with lower variance than the single models that comprise it. Bias-variance tradeoff haunts supervised learning algorithms. A model's expected generalization error depends on the sum of bias and variance. Bias is intrinsic to the learning algorithm, but variance may come from several sources. In our case, the bias is fixed since we do not change the FHMM algorithm, but the variance may be from data insufficiency or overfitting. Averaging model parameters in the perceptron alleviates overfitting. But data insufficiency still contributes to variance, which is the situation for active learning in the early stages. Creating ensembles is an effective means for reducing variance. But reducing member variance is not equal to reducing ensemble variance. Ensemble variance is what we care about ultimately, since it directly relates to the generalization of the ensemble on unseen data. The actively selected examples might be used to reduce single model variance, which might not significantly help reducing ensemble variance. Thus, iterations might be wasted, and active learning slows. We employed a new example sampling method that forces the five FHMM members to be more "similar" to each other in early iterations, to reduce the selection of edge case examples that could cause variance of the ensemble. The technique is named "sentence re-weighting", introduced below.

We introduce the general architecture of a typical active learning routine for ensembles [22] and [23] before we discuss details of active learning. The general framework has been adapted to sequence labeling problems (Algorithm 2).

In this work, we configured algorithm 2 by using (1) a sequence labeling utility function *Sequence Vote Entropy* [36] for *Utility*(), and (2) a new *Sentence Re-weighting* method for creating ensembles from *T*, which corresponds to *learn_Ensemble* (Base, T) shown in Algorithm 3.

*Utility Function.* The purpose of a utility function is to evaluate how much the classifier would gain from the unlabeled training example. To do this, we applied the utility function *Sequence Vote Entropy* (**SVE**). We are interested in this method because it evaluates the probability for the whole sequence instead of evaluating each token individually. We did not compare exhaustively all potential utility functions since this is not the purpose of the paper. We used sentence random sampling to compare against it. **SVE** utility is calculated as

$$\phi^{SVE} = -\sum_{\hat{y}\in N^c} P(\hat{y}|x; M) \log P(\hat{y}|x; M)$$

Where $N^c$ is the collection of all *N*-best predictions from models in the ensemble. Here, $x$ is the input sentence; $\hat{y}$ represents a prediction from $N^c$, which is the collection of all n-best predictions from models in the ensemble $M$. Each sequence prediction probability is calculated as the average of all the prediction probabilities for each model $m$ by $P(\hat{y}|x; M) = \frac{1}{c}\sum_{m=1}^{M} P(\hat{y}\,|x; m)$, and each probability within the sum is the sequence probability calculated by the *Viterbi Decoding* algorithm. Also, the probability has been normalized into [0,1] interval being divided by the sum of probabilities.

*Sentence Re-weighting.* Once an example *e* is chosen by the utility function, it will be added to *T*, the collection of annotated training data. It will be sampled with equal probability as a normal sample in *T*, as in the general ensemble creation algorithm. Imagine the case where *e* is only involved in a small portion of FHMM members in ensemble. For a sentence that is extremely similar to *e*, say, **e'** in the unlabeled example pool, it might not need to be trained again since *e* is included in *T*. But, because *e* was only included in some of the individual FHMM model members, some of the models will not be able to predict the similar instance **e'** correctly, just because *e* is not included in those models. So some members of the ensemble will disagree on predicting **e'**, which will lead to a high utility value of **e'** that makes it likely to be selected again in the active learning process. Since **e'** is highly similar to e, the annotation and training of **e'** is redundant and slows active learning. In order to reduce the selection of redundant examples like **e'**, we created a strategy to re-weight the examples in the labeled training data that gives newly labeled examples a higher probability of being sampled in the very next iteration. And then the chance of newly-labeled samples being selected will decrease over iterations. This is analogous to short-term memory mechanism used in Long-Short Term Memory (LSTM) neural networks [12].

In algorithm 2, step 2, *learn_Ensemble*() randomly samples with equal probability each example in *T* with expectation *r*, called the *sample rate*. $\mathbb{E}(|T_m|) = rT$ is the expected number of examples fed to model *m*. We fix the rate *r*, and in iteration *t*, we choose the top example *i* (with highest utility), and set the probability to 1 for the example in next (*t*+1)-th iteration. This means the newly added example will always be included to build the ensemble in iteration *t*+1, for every FHMM member. Then in (*t*+2)-th, the probability for this instance will be max $(r, w_{it})$, where $w_{it}$ is the weight of example *i* at iteration *t*. The weight decay for example *i* will stop until $w_{it}$ reduces to *r*. See Algorithm 3. For any example-weight pair in training data $T$ of time $t$, we first modify the weight $W_{it}$ for the example $T_{it}$ by deducting decay factor $\alpha = 2(1-r)/(t-1)$, which is calculated by equating two expectations

$$\mathbb{E}(|T_{t+1}|) = r(t+1)$$

Which is the expected number of examples selected by uniform random sampling, and the new expectation

$$\mathbb{E}'(|T_t + 1|) = t - 0.5\alpha t(t-1) + rI$$

Which is the expected number of examples selected by sentence reweighting. Equating the two guarantees that the sentence re-weighting method only changes the weights of the examples in $T$ instead of the total number of training examples. The $\alpha$ value used to deduct from the probability decreases through iterations. In other words, the rate of weight change will decrease over time. When the weight decreases to a pre-defined rate $r$, it will stop decreasing, and the corresponding example $T_{it}$ will no longer have a high sampling probability. In the Experiments section, we will see the effect of this example re-weighting strategy.

## 7. EXPERIMENTS

(1) We compare the FHMM model to others on the basis of similar sequence-labeling data sets—a necessity since ours is the only model that is used to handle EST data (2) we consider the active learning process, and (3) the degree to which the inference algorithm affects ensemble FHMM, and (4) we run our ensemble FHMM model on our data set to check precision and recall separately on time expressions and on location expressions.

### 7.1 Corpora to compare models

We chose the following datasets to compare our Perceptron FHMM with several other models. Due to data annotation and others' published results, we compared on the basis of NER and the chunking task. We use the CoNLL 2000 Chunking dataset, the NLPBA data set for Biomedical NER, and OntoNotes version 5.0 – the 5 news subsets ABC, MNB, NBC, PRI and VOA therein. We skipped the CoNLL 2003 NER dataset because at the time of the paper we are not able to acquire the dataset. We would leave for our future work. We can use part of it, but it seems unnecessary. PRI and VOA are two datasets that resemble EST dataset in size. ABC, MNB and NBC are three small NER datasets that simulate early stages in active learning. NLPBA is used for testing biomedical NER and CoNLL chunking is used for testing a NER-like task.

### 7.2 Perceptron FHMM vs. previous work

Our task of finding spatiotemporal phrases is essentially the recognition of non-named entities. However, in order to compare our model to others, we compare on task of Named Entity Recognition, and the F1 metric that is quoted in others' research. We look also at the chunking task that recognizes phrases because of its similarity to recognizing spatiotemporal phrases. We use the F1 measure from our single FHMM model to be consistent with others' published results on single models (even though [37] uses an ensemble HMM and voting to achieve 95.23% on the chunking task, which was the highest reported). For training the FHMM, we stops training iterations based on whichever is met first: 100 iterations or the 1E-10 training error rate. The decoding algorithm uses Viterbi algorithm. The features used by FHMM model on each dataset were described in Table 4.

Table 6 shows results on the CoNLL 2000 chunking task. Table 7 shows the F1 results for the OntoNotes news dataset. Each dataset was split into about 80% of the data for training and about 20% for testing. Train test split of those follows [8] to be comparable. Our score is higher than that of Finkel and Manning's [8] basic NER model which uses features similar to those that we use. Our score is lower than the joint model that uses parsing information. [29] used Average Perceptron training along with rich semantic features, as do [41].

Table 8 shows how different models perform the NER task on the NLPBA dataset. The NLPBA 2004 dataset is an English language dataset for biomedical NER, which includes PubMed abstracts along with corresponding named entities (such as DNA and RNA). This data differs from the OntoNotes news articles in that the terms are biomedical terms with a high out-of-vocabulary rate and with complicated word forms. Still, it is the Wang et al [49] ensemble method that outperforms other single-model methods to a large extent. The second best result comes from using rich knowledge as features for word matching. Among the results that do not use extended outside knowledge but explore word forms and context only, FHMM is shown to be competitive with [8] and [35]. Tables 7 and 8 suggest that our FHMM model needs improvement to complete with knowledge rich models or the joint model with parsing information. This suggests that additional information helps.

Figure 3 shows how the *best Viterbi sequence* (BVS) compares to the *Best BP sequence* (BPS) on data sets of different sizes, and how different data sampling methods compare for active learning. Experiments are conducted both on the OntoNotes corpora and on our own Expanded Spatiotemporal Dataset (see Table 5). We used an ensemble of 5 FHMMs, with a sample rate $r$=0.8, and compared results with **vt** (Viterbi Decoding), **bp** (Belief Propagation), **nrw** (no example re-weighting for creating ensembles), **rw** (example re-weighting), **utl** (utility function to select unlabeled examples), and **rnd** (random sampling from unlabeled examples--baseline).

Table 5 Data set size for model comparison

| Data set | Train | Test | Data set | Train | Test |
|---|---|---|---|---|---|
| OntoNotes ABC | 971 | 194 | CoNLL Chunking | 18456 | 3856 |
| OntoNotes MNB | 425 | 212 | NLPBA | 8936 | 2012 |
| OntoNotes NBC | 506 | 135 | EST Dataset | 1701 | 450 |
| OntoNotes PRI | 1648 | 401 | | | |
| OntoNotes VOA | 1500 | 397 | | | |

Table 6. Model Comparison on CoNLL 2000 Chunking

| Model | F1 |
|---|---|
| FHMM Single model- Viterbi | **94.48** |
| Conditional Random Fields [34] | 94.3 |
| Latent Dynamic Conditional Random Fields [38] | 94.34 |

Table 7. F1 Model comparison on OntoNotes NER Datasets

| Model | ABC | MNB | NBC | PRI | VOA |
|---|---|---|---|---|---|
| FHMM Single model-Viterbi | 74.87 | 68.9 | **68.59** | 81.5 | 83.14 |
| CRF by Finkel et al. [8] | 74.91 | 66.49 | 67.96 | **86.34** | **88.18** |
| Average Perceptron by Ratinov [29] | 72.74 | **73.1** | 65.78 | 79.63 | 84.93 |
| CRF by Tkachenko [41] | **76.45** | 71.52 | 67.41 | 83.72 | 87.12 |

Table 8. F1 Model comparison on NLPBA Dataset

| Model | F1 |
|---|---|
| FHMM | 71.42 |
| Ensemble of Sequence Labeling algorithms [48] | 77.6 |
| Average Perceptron with knowledge [41] | 74.27 |
| CRF by Wang et al. [49] | 70.1 |
| CRF by Settles et al. [35] | 69.8 |

(1) *Viterbi* **(vt)** vs. *Belief Propagation* **(bp)**: Comparing the curves for **vt** and **bp** in Figure 3, we can see that the BP algorithm attains higher overall accuracy on almost all datasets to a large margin, and the differences amplify when the iterations increase. We conducted 250 iterations for each dataset, because performance increases slowly afterwards. The difference between using BVS and BPS for inference seems to have roughly 10% F1 difference on our EST data. The phenomenon can be explained by the discussion in section 4. To review, for BVS, we simply aggregate the sequences from each model output, and the sequence from each model is used for voting. By contrast, BPS votes each token individually, so the method has the ability to modify part of the sequence (that is, part of the sentence) when giving outputs. This is an important observation because when we conduct active learning, the data size is small for the first several iterations, and we need a reliable decoding algorithm that tells us how uncertain is each training instance, so that utility values can be computed more reliably based on those predictions.

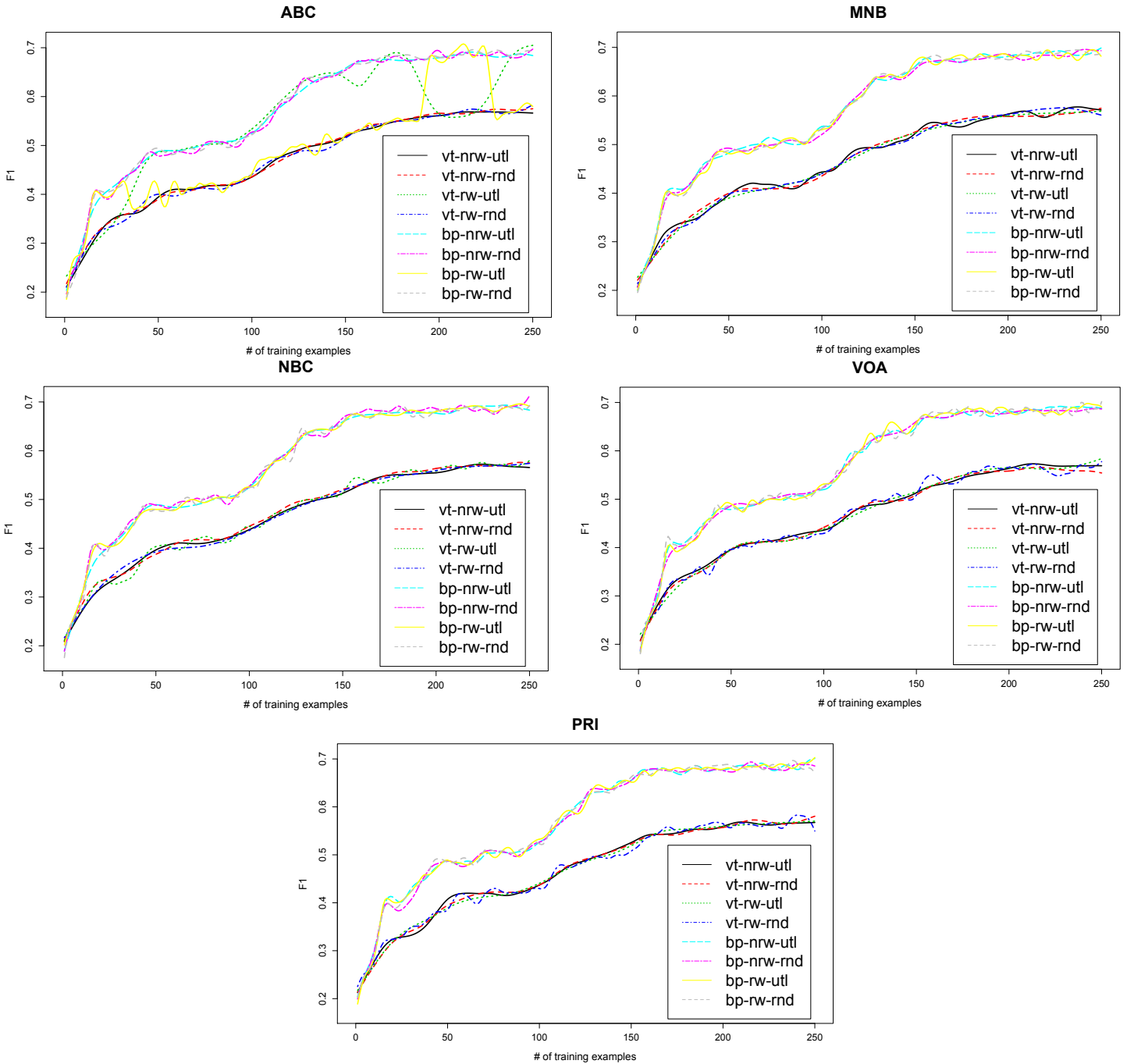

**Figure 3 Comparison of sampling strategies for ensembles during active learning process**. ABC, MNB, NBC, PRI, VOA are 5 subsets of the OntoNotes news corpus. Each dataset is split into roughly 80% train 20% test. x-axis is the number of training examples labeled so far, and the y-axis is the F1 value on each corresponding test dataset. We used an ensemble of 5 FHMMs, sample rate $r$=0.8.

(2) **With Utility Functions (utl) vs. Without (rnd).** From the Figure 3, by comparing curves with **utl** and **rnd** to the other parameters fixed, we can see that using a utility function did not significantly outperform the random sampling baseline. This result is consistent with the [36] result. This proves that for sequence labeling in active learning, choosing the whole sequence performs a kind of smoothing for the learning curve, because the non-informative tokens in the sequence have to be labeled with the informative ones. Better solution could be selecting part of sentences to label instead of the sentence as a whole.

(3) **With *Example Re-weighting (rw)* vs. without (nrw)**. From the comparison between the curves with 'rw' and 'nrw', we can see that the re-weighting method was slightly better than the non–re-weighting method especially on our EST dataset (see the two entangled curves at the bottom of Fig. 4). This confirms our hypothesis that the short-term memory provided by the weighing scores *does* help ensembles remember previous states, and make better choices on the example to learn in the next iteration. Note that in the five OntoNotes datasets, this pattern does not show clearly because of the vibration of 'rw' curves.

We mentioned above that we used two decoding algorithms for inference, *best Viterbi sequence* and *best BP sequence*. In Table 9, we experimented with two inference algorithms using the OntoNotes dataset, which resembles our EST dataset with respect to data size and number of labels (although the EST has a wider variety of label). The *Best BP sequence* outperformed the *best Viterbi sequence* by a small margin, and we observed a similar discrepancy in our experiment with the EST dataset, as Figure 4 shows.

We saw inaccuracies in Figure 3 that prevented steady improvement in accuracy as training data size increases, because active learning introduces random errors. Errors come from noise in the training data, and from absorbing entire sentences for training regardless of the utility of the single tokens. We could improve active learning by not labeling tokens that are less informative, and labelling only the informative parts of a sentence.

**Table 9. Comparison of two decoding algorithms on OntoNotes NER Dataset**. An ensemble of 5 FHMMs is used. Each FHMM uses 80% of randomly sampled examples

| Model | ABC | MNB | NBC | PRI | VOA |
|---|---|---|---|---|---|
| FHMM Ensemble –Viterbi | **73.74** | 66.7 | 65.62 | 78.33 | 81.67 |
| FHMM Ensemble - BP | 70.41 | **69.0** | **68.09** | **81.56** | **82.93** |

## 7.3 Evaluation of spatiotemporal parsing

The purpose is to evaluate how well our ensemble model does on our defined **G** and **T** tags. In Table 10, we used 1700 training sentences on the 5-model ensemble, at a sample rate of 0.8. During preliminary experimentation, we had created *79 spatiotemporal sentences* to determine which types of temporal expressions temporal parsers found. These 79 sentences do not contain named entity tags, which can be extremely helpful to non-entity spatiotemporal expressions, because entities co-occur with non- entity spatiotemporal expressions frequently. So we created an FHMM ensemble model on the EST dataset with all tags except for T and G tags, namely **FHMM-G,** to train a general purpose NER system on location, dates, organization, person, etc.

Then, FHMM-G is applied to features relevant for the corpus (as shown in Table 4).

Our model did remarkably well on the precision (89.27) and recall (88.17) for temporal expressions. Naturally, it did less well on location expressions since there were very few in the 79-sentence corpus.

**Table 10. F1 of FHMM on spatiotemporal expressions. G T means expanded tags. 'Entities' means all but G &T.**

|  | Geospatial (G) | | | Temporal (T) | | | Entities |
|---|---|---|---|---|---|---|---|
|  | P | R | F1 | P | R | F1 | F1 |
| EST Test | 87.7 | 86.85 | 87.27 | 90.71 | 92.18 | 91.43 | 89.06 |
| 79 Sentences | 87.23 | 79.24 | 83.04 | 89.27 | 88.17 | 88.71 | 88.6 |

Our hypothesis that joint modeling would be effective is based on the large number of linguistic parallels between time and place language (as shown in Table 3), which we were able to disambiguate by constructing a model that parses expressions both for location and also for time. The fact that we achieved high recall and also precision for both location and temporal phrases (see Table 10) demonstrates that our joint spatiotemporal parser is successful.

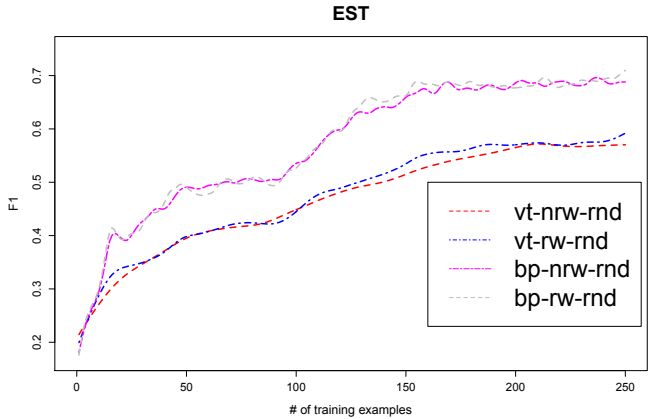

**Figure 4 Comparison of Decoding methods and sentence re-weighting methods for EST dataset. No utility function used.**

## 7.4 Generalizability and future research

The algorithm is limited in that it was trained using certain seeds. Spatiotemporal expressions not included in the data set still will not be found, nor will numerical sequences with numbers different than those in the training data. We could widen the expressions recognized by using templates for features. For example, we could find "the x week of y", rather than only "the second week of April". Even so, a larger quantity of data would need to be annotated manually, by crowd-sourcing on Mechanical Turk, for example, so that the patterns would be tagged.

Alternative algorithms for training that could be tested include neural networks, which have shown excellent results on sequential labeling tasks, especially the long-term short-term memory network [12]. The method lends itself in particular because the sequences that we would recognize are longer than standard Named Entity Recognition tasks.

## 8. CONCLUSION

We have identified the spatiotemporal expressions missed by state-of-the-art time and location parsers, and have created a tool that will identify such expressions. The recall of our spatiotemporal parser is high, and does not sacrifice precision, which demonstrates the strength of our solution.

To create our spatiotemporal parser, we collected and annotated a data set rich in such expressions for training the model. We used active learning to reduce the manual annotation required and chose only the most effective examples for training. We used an average perceptron FHMM as a base model to recognize such expressions and we adopted ensemble learning to create a bag of FHMMs to reduce the model variance. Our dataset is annotated both with named entities and with unnamed entities that are spatiotemporal expressions is available to the research community. Our approach can be applied widely in sequence labeling tasks.

We have shown that ensemble methods are effective in reducing the variance of label output. Further, we have found surprisingly that Belief Propagation is a more suitable inference algorithm than Viterbi decoding at the beginning of the active learning periods, owing to the small size of the data. Our re-weighting for active learning sampling outperformed the baseline random sampling, but the utility function does not seem to improve active learning very much.